# Chronic pain patient narratives allow for the estimation of current pain intensity

**Diogo Afonso Pedro Nunes – corresponding author**
*INESC-ID, Lisbon, Portugal*
*Instituto Superior Técnico, Universidade de Lisboa, Lisbon, Portugal*
diogo.p.nunes@inesc-id.pt
0000-0002-6614-8556

**Joana Maria de Pinho Ferreira Gomes**
*Department of Biomedicine, Experimental Biology Unit, Centre for Medical Research (CIM), Faculty of Medicine, University of Porto, Porto, Portugal*
*i3S - Instituto de Investigação e Inovação em Saúde, University of Porto, Porto, Portugal*
jogomes@med.up.pt
0000-0001-8498-6927

**Daniela Santos Oliveira**
*Rheumatology Department, Centro Hospitalar Universitário São João, Porto, Portugal*
*Medicine Department, Faculty of Medicine, University of Porto, Porto, Portugal*
*Center for Health Technology and Services Research (CINTESIS), Faculty of Medicine, University of Porto, Porto, Portugal*
danielasoff@gmail.com

**Carlos Jorge Cabral Vaz**
*Rheumatology Department, Centro Hospitalar Universitário São João, Porto, Portugal*
*Medicine Department, Faculty of Medicine, University of Porto, Porto, Portugal*
*Center for Health Technology and Services Research (CINTESIS), Faculty of Medicine, University of Porto, Porto, Portugal*
carlosjcvaz@gmail.com

**Sofia Santos Pimenta Vale**
*Rheumatology Department, Centro Hospitalar Universitário São João, Porto, Portugal,*
*Medicine Department, Faculty of Medicine, University of Porto, Porto, Portugal*
sofiadsp@sapo.pt

**Fani Lourença Moreira Neto**
*Department of Biomedicine, Experimental Biology Unit, Centre for Medical Research (CIM), Faculty of Medicine, University of Porto, Porto, Portugal*
*i3S - Instituto de Investigação e Inovação em Saúde, University of Porto, Porto, Portugal*
fanineto@med.up.pt
0000-0002-2352-3336

**David Manuel Martins de Matos**
*INESC-ID, Lisbon, Portugal*
*Instituto Superior Técnico, Universidade de Lisboa, Lisbon, Portugal*
david.matos@inesc-id.pt
0000-0001-8631-2870

## Article information
Number of figures: 6
Number of tables: 7

# Chronic pain patient narratives allow for the estimation of current pain intensity

## Abstract (250 words)

Chronic pain is a multi-dimensional experience, and pain intensity plays an important part, impacting the patient's emotional balance, psychology, and behaviour. Standard self-reporting tools, such as the Visual Analogue Scale for pain, fail to capture this burden. Moreover, this type of tools is susceptible to a degree of subjectivity, dependent on the patient's clear understanding of how to use it, social biases, and their ability to translate a complex experience to a scale. To overcome these and other self-reporting challenges, pain intensity estimation has been previously studied based on facial expressions, electroencephalograms, brain imaging, and autonomic features. However, to the best of our knowledge, it has never been attempted to base this estimation on the patient narratives of the personal experience of chronic pain, which is what we propose in this work. Indeed, in the clinical assessment and management of chronic pain, verbal communication is essential to convey information to physicians that would otherwise not be easily accessible through standard reporting tools, since language, sociocultural, and psychosocial variables are intertwined. We show that language features from patient narratives indeed convey information relevant for pain intensity estimation, and that our computational models can take advantage of that. Specifically, our results show that patients with mild pain focus more on the use of verbs, whilst moderate and severe pain patients focus on adverbs, and nouns and adjectives, respectively, and that these differences allow for the distinction between these three pain classes.

## 1. Introduction (500 words)

Language is a key communicator for the task of clinical chronic pain assessment and management [32,19]. A narrative of the personal experience often includes valuable information about the bodily distribution of the feeling of pain, temporal patterns of activity, intensity, emotional and psychological impacts, and others. Additionally, the choice of words may reflect the underlying mechanisms of the causal agent(s) [32], which in turn can be used by the physician to redirect therapeutic processes. This sub-language has been studied in previous research, such as the structuring of the Grammar of Pain [7], and the study of its lexical profile, which resulted in the McGill Pain Questionnaire (MPQ) [18] that is widely used to characterize pain from a verbal standpoint in clinical settings [12,27].

There is a special focus in estimating the pain intensity of patients currently in pain, based on a variety of features, because subjective reports, such as reporting pain intensity through a Visual

Analogue Scale, are susceptible to inconsistencies, impression or deception biases, socially suggestive cues, and others [8]. Most research works are based on facial expressions [8,3,33,29,11,34], but there have also been successful works based on electroencephalograms [13], brain imaging [14,17], and autonomic features [15]. To the best of our knowledge, there is no published research work that attempts to estimate current pain intensity from language, specifically from chronic pain patient narratives.

Other works have focused on computationally studying the language of pain, providing evidence for the valuable information conveyed by this sub-language, though this was not done to directly estimate pain intensity. Tan et al. [28] explored what people report on social media about their experience of chronic pain to automatically update MPQ descriptors. Foufi et al. [6] developed a pipeline to extract and interpret biomedical entities and relations, also from social media posts. In the same social media line, we have also proposed in a previous unpublished work (21) a method to model the language of chronic pain, as found on the Reddit platform, from differing base-pathologies, and to compare these models qualitatively and quantitatively. However, not many works have attempted to use patient narratives directly obtained from registered chronic pain patients, thus, lacking other demographic and clinical parameters that are essential to contextualize their results. This is probably due to the costs associated with designing and implementing a set of experiments with human subjects around their health. Berger et al. [4] analyzed language features, directly obtained from patient narratives, to predict chronic pain placebo responders, with 79% accuracy. In a previous work [22], we also used patient narratives to classify chronic pain patients to one of two rheumatology base-pathologies, also with 79% accuracy. These works highlight the importance of the linguistic expression of chronic pain for the computational assessment.

In this work, we raise the hypothesis that Natural Language Processing (NLP) techniques can extract useful information from chronic pain patient narratives, and that this information can be used to estimate the current pain intensity and better understand the language of chronic pain.

## 2. Materials and methods (no word limit)

### 2.1. Materials

### 2.1.1. Participants

A total of 94 chronic pain patients participated in this study. These are adults (older than or equal to 18 years of age), of either sex, diagnosed with osteoarthritis, rheumatoid arthritis, or spondyloarthritis (including psoriatic arthritis) for at least three months. All participants had to have had reported pain symptoms for at least three months.

### 2.1.2. Data privacy

All data were collected under a collection protocol (number 304/19), approved by the Ethics Commission of the University Hospital Center of São João (CHUSJ), in which data confidentiality is explicitly protected. All participants signed an informed consent form before data collection took place. Participants were made aware of the study's purpose before giving consent. All data are anonymous, and the presentation of results is always made without individual references. Participant data were identified with a unique ID, and kept separate from the ID resolution key, which is maintained in physical format at a secure location. Participant personal identification is never used.

### *2.1.3. Data collection protocol*

Data collection took place at the Rheumatology department of CHUSJ from October 2019 to October 2020. At the end of the regularly scheduled appointment with the pain clinicians, participants were informed of the study, its objectives, and of which data would be collected. This included, in order, the verbal description of the chronic pain experience (i.e., patient narrative resulting from the interview, recorded with a smartphone, exclusively used for this purpose) and additional contextual demographic and clinical information. After giving their informed consent, the data collection process started. The whole collection process was performed by the clinician, in Portuguese, since all participants were fluent in that language.

Since language, sociocultural, and psychosocial variables are intertwined, the context in which the patient is describing the experience of chronic pain affects both the vocabulary and how it is reported [9]. Thus, data collection took place after the scheduled appointment with the (already familiar) clinician, in the appropriate clinical context. Moreover, the interview was performed before collecting demographic and clinical data so as to minimize any biasing of the patient and mimic the predictive nature of our task.

### *2.1.4. Patient interview*

The interview that each patient responded to is shown in Table 1, as well as the topic that each question focuses on, and the reasoning behind it. Even though there was no test pilot with actual patients to investigate possible limitations of this interview script (due to time constraints), the questions were designed in conjunction with two pain clinicians, from the same Rheumatology department, who have multiple years of experience in communicating with chronic pain patients, and in clinically assessing and managing them.

Contrarily to other works (e.g., Berger et al. [4]), this interview does not include warm-up questions, nor more than seven questions, to limit the impact on the appointment schedule of the clinicians to no more than 5 minutes after the scheduled time. Moreover, since the data collection was performed by the clinician that just participated in the regular appointment, the participant was already comfortable talking with that person. Also, the recording device was chosen to be a smartphone, exclusively used for this purpose, as already stated, laying on the table between the clinician and the participant, to cause minimal visual impact. To maximize the number of participants in the restricted time interval of data collection, three clinicians participated in that step of the study. As stated before, psychosocial variables might affect chronic pain perception and description, including who the interviewer was, how the interviewer was presented, and how the interviewer talked to the patient [31]. For these reasons, we also consider the interviewer to be an independent variable.

### *2.1.5. Demographic and clinical data*

Besides the interview, we also collected additional data that consisted of basic demographic information, clinical parameters deemed complementary, and the self-reported intensity of pain. Demographic data consisted of age, gender, highest completed education, current (or last) professional occupation, and whether the participant was professionally active, or not. Complementary clinical data consisted of the number of years since the initial diagnostic, number of years since the first reports of pain, medication, values of Erythrocyte Sedimentation Rate (ESR), and C-Reactive Protein (CRP). Finally, pain intensity and disease state were self-reported using a Visual Analogue Scale (VAS). Even though these were reported on a continuous scale, they were discretized afterwards into three classes: mild (1-4), moderate (5-6), severe (7-10). We decided to discretize these reports due to the limited number of participants. These pain intensity categories have been suggested by Serlin et al. [25], the results of which other studies have been able to replicate [1]. This categorization does not implicate loss of relevance to the clinical practice since it is already commonly used between pain practitioners and patients [1].

*2.2. Methods*

*2.2.1. Participant data preprocessing*

Each interview recording file was semi-automatically segmented into clinician and patient speech audio segments, and manually labelled with the corresponding question of the interview. Each patient speech segment was manually transcribed, maintaining all speech disfluencies that were expected to occur, such as repetition, correction, and syntactic errors [26]. After transcription, no meta-annotations were made, maintaining only the exact words the patient uttered (excepting filling sounds, such as "uhm" and "uh", which were not transcribed). This method was chosen to keep the textual data as close as possible to the natural utterance of the patient. However, punctuation had to be inferred from the pauses and natural flow of the sentences. Finally, patient text was automatically Part-of-Speech (POS) tagged and lemmatized using STRING [16]. In this case, POS tagging encompasses assigning one of the POS tags used by STRING to each token (i.e., occurrence of a word) in the text (STRING uses a total of 13 tags, which denote a noun, verb, adjective, determinant, pronoun, article, adverb, preposition, conjunction, numeral, interjection, past participle, and relation). Lemmatization encompasses replacing each token by its root-form, also called lemma (e.g., "hurt" is the root-form of "hurting"). These are common text normalization steps which allow for textual noise reduction, without losing the core syntax and semantics of what the patient is trying to convey. Complementary demographic and clinical data were collected using a physical form, filled out by the clinician performing the interview, except for the VAS, which was marked by the participant. Each form was manually transferred to a row in a spreadsheet. Fields that were left empty by mistake, or because the patient did not want to answer, were left empty in the spreadsheet. These processes were performed and validated by the authors and were sampled for possible mistakes and corrections.

*2.2.2. Confounding variables on pain intensity*

Since the objective of this work is to use language to estimate current pain intensity, all other variables must be controlled, such as age, pathology, years in pain, and so on. Due to the limited availability of participants, these could not be controlled a priori, nor during experimentation, excepting the hospital unit (rheumatology) from which participants were recruited. Therefore, to maintain internal validity, demographic and clinical variables (excepting pain and disease VAS) were analysed in order to understand to which extent they can explain pain intensity differences in our population. As previously mentioned, we also included the interviewer as an independent variable in this analysis.

### 2.2.3. Language features extraction

Each participant's interview was used to extract a set of features that characterize that participant. These features are: basic verbosity (3 features: word count, interview length, and word rate), Term Frequency / Inverse Document Frequency (TF-IDF) on the lemmatized text concatenated with the corresponding POS tag (number of features equal to the size of the shared vocabulary), TF-IDF on the POS tags of each lemma (13 features, each corresponding to one of the tags), and, finally, topic distribution (12 features, each corresponding to the weight given to each of the topics automatically found in the corpus, using CluWords [30] – this number was obtained in previous experiments with the same data (20)).

The motivation behind the choice of these features is as follows. Basic verbosity features function as a proxy to the capability and willingness to describe the personal experience of chronic pain. Our hypothesis is that the less the patient has to say in general, the less impactful the experience of pain is in the quotidian life.

TF-IDF on the lemmatized text (concatenated with the corresponding POS tag) is a set of vocabulary-based features that aims at distinguishing patients based on word choice and frequency, approximating the underlying semantics of the narrative. To this end, we removed common words that do not convey significant semantics (using the standard list of Portuguese stop-words from the Natural Language Tool Kit (NLTK) [5]). Our hypothesis is that the frequent (or infrequent) choice of specific words to describe the experience of chronic pain correlates with specific thresholds of pain intensity.

Similarly, TF-IDF on the POS tags of each lemma is a set of syntax-based features that aims at distinguishing patients based on syntactic constructions and their frequency. Our hypothesis is that the preference for certain syntactic categories correlates with certain aspects of the experience of chronic pain (e.g., verbs may correlate with actions), and their frequency may be indicative of pain intensity thresholds.

Finally, topic distribution is a set features that aims at capturing high-level semantic concepts and their relative importance to each participant. Each question of the interview was designed to focus on specific topics of the experience of chronic pain. Our hypothesis is that whether those topics are important or not to the patient is reflected in how much that patient discusses those topics (topic weight). As we previously mentioned, pain research literature has shown the correlation between these topics and the perceived experience of chronic pain.

### 2.2.4. Experimental setup for pain intensity estimation

Our task is formally defined as a multi-class classification problem. We tested the performance of two machine learning classifiers: the Decision Tree Classifier (DT), and the Support Vector Machine (SVM) (with the implementation and default parameters from the Sci-kit Learn package [23]). These models have been extensively used in similar works [8,13,14]. We measured model performance according to the weighted $F_1$ score. This metric takes the weighted average of the $F_1$ scores of each class, where the class weights are inversely proportional to their prevalence in the dataset. We purposefully focused the evaluation on the $F_1$ score and not others, such as accuracy, precision, or recall, because it already provides an aggregate view of the last two (equally weighted), and accuracy is not well-suited for imbalanced multi-class settings. Formally, the weighted $F_1$ score is given by

$$\frac{1}{\sum_{l \in L}|y_l|} \sum_{l \in L} |y_l| F_1(y_l, \hat{y}_l)$$

where $y$ is the set of true labels, $\hat{y}$ is the set of predicted labels, $L$ is the set of labels, and $y_l$ is the subset of $y$ with label $l \in L$.

We chose these models due to their ease of use, interpretability, and low complexity, which is very relevant for a limited sample size, such as our case. For the same reason, we trained and validated our machine learning models in a Leave One Out Validation (LOOV) setting, which, again, is extensively used in similar works [3, 33, 29]. In this setting, there are $n$ distinct training partitions of $n – 1$ samples, and $n$ distinct validation partitions of 1 sample. Importantly, the one-out sample is never used for feature selection nor model fitting.

As a baseline, we used the Zero Rate Baseline (ZRB) classifier, which deterministically predicts the most prevalent target class in the training set. This is an important comparison for imbalanced datasets because it is relatively easy to obtain high performance scores when one class is much more prevalent than the other(s). Thus, a classifier cannot be deemed useful if it performs worse than, or as good as, the ZRB.

We tested the DT and SVM classifiers with 6 different sets of features, effectively obtaining 12 distinct models. The sets of features are: (1) verbosity, (2) TF-IDF, (3) POS TF-IDF, (4) topic distribution, (5) Early Fusion (EF), and (6) Late Fusion (LF). In the case of EF [24], all features were concatenated into a single feature-vector before inputting to the machine learning model. In the case of LF [24], a different model was fitted with each set of features, predicting the target class of a given sample using a majority voting scheme [2] between the predictions of the different models. By definition, the ZRB performs equally for all feature sets, since it ignores them.

Importantly, given one of the 6 sets of features to perform pain intensity estimation, we used only the top-$k$ features, based on the SelectKBest feature selection method provided by Sci-kit Learn. This ranking is given by their ANOVA F-value statistical significance, and $k$ ($k \in [2,20]$) is given by the parameter search GridSearchCV method, from the same package. The number of features was limited to a maximum of 20 to reduce the feature space due to the limited number of samples. Features with zero variance were also discarded.

### *2.2.5. Influence of confounding variables on language*

Language is the observed variable of our study, from which we extract features and infer the pain intensity category (by this definition, the dependent variable). However, language itself may depend on a myriad of sociocultural, demographic, and even clinical parameters. As stated before, these parameters could not be controlled, due to the availability of participants and time constraints. Nonetheless, we wanted to account for this dependency and modulation of language, which can possibly affect the prediction task. To this end, for each demographic and clinical parameter, we discarded all language features that lead us to reject the null hypothesis of independence with that parameter (for categorical parameters we use one-way ANOVA with the standard threshold of $p < .05$, and for continuous parameters we use Pearson's r with the threshold $|r| > .7$). We compared the best performing model from the previous section (which ignores all language dependencies) with the model trained on the subset of the remaining features. This scenario acknowledges the possibility that language is being modulated by certain demographic and/or clinical parameters, and estimates pain intensity using only those language features which fail to reject the null hypothesis of independence.

### *2.2.6. Classifier decision interpretation*

Taking the best scoring model and set of features, we used this information to extract insights regarding the linguistic differences between mild, moderate, and severe pain, when describing the personal experience of chronic pain.

## 4. Results (no word limit)

### *4.1. Participants*

A total of 94 participants were enrolled in this study. Of these, 65 completed the full interview and have data for all complementary data fields. The remaining participants are not accounted for in the rest of this study.

In terms of demographics, the study population is composed of 25 males ($58.3 \pm 10.6$ years of age) and 40 females ($55.2 \pm 13.8$ years of age). Overall, participants are $56.4 \pm 12.7$ years old. The highest education level of 49 participants was the basic level, whilst for the remaining 16 was the high-school level. 34 participants were professionally active, and the remaining 31 were retired or on sick leave at the time of data collection. In terms of clinical state, 32 participants were diagnosed with spondyloarthritis, 29 with rheumatoid arthritis, 2 with psoriatic arthritis, and 5 with osteoarthritis. Importantly, 2 participants had comorbid osteoarthritis and rheumatoid arthritis, and 1 participant had comorbid osteoarthritis and spondyloarthritis. Participants had the diagnosis, on average, for $12.4 \pm 9.5$ years, whilst have reported pain, on average, for $15.9 \pm 11.0$ years. Participants were interviewed by 3 clinicians. Of the 65 considered participants, each clinician performed 50, 9, and 6 interviews, respectively. Finally, regarding pain intensity category, 38 participants reported mild pain, 12 moderate pain, and 15 severe pain, as inferred from the discretized VAS data. Table 2 tabulates these distributions per pain intensity category.

### *4.2. Quality control of interview preprocessing*

From the list of 65 participants, we sampled 10 participants, at random, to assess the quality of the preprocessing steps, which encompass automatic, semi-automatic, and manual processes. The sample included 86 audio segments, each manually transcribed to an individual plain text file, of which 7 included at least one transcription error of some sort (8.2% error rate). However, in 4 of those cases, the error was due to inaudible words, which still could not be transcribed. Finally, the sample included 1914 POS-tagged and lemmatized tokens, of which 46 were incorrectly tagged or lemmatized (2.4% error rate). Given these results, we deemed unnecessary to check all transcription, POS-tagging, and lemmatization events for possible errors. All errors found during quality control were corrected (when possible).

*4.3. Influence of confounding variables on pain intensity*

Starting with demographic variables, no statistically significant differences were observed in the three pain intensity categories regarding age (one-way ANOVA; mild, 55.8 ± 13.0; moderate, 54.7 ± 12.5; severe, 59.1 ± 10.9; $p$ = .63). However, a chi-square test for independence led us to reject the null hypothesis that pain intensity and gender ($p$ < .001), education level ($p$ < .001), and whether the participant is professionally active or not ($p$ = .002), are independent variables. Importantly, all pain intensity categories have similar percentages of professionally active and inactive participants. Regarding clinical variables, no statistically significant differences were observed in the three pain intensity categories regarding the number of years since the initial diagnostics (one-way ANOVA; mild, 11.3 ± 9.5; moderate, 11.9 ± 9.1; severe, 15.4 ± 8.6; $p$ = .37), the number of years since the first reports of pain (one-way ANOVA; mild 14.6 ± 11.2; moderate, 14.8 ± 10.7; severe, 20.1 ± 9.0; $p$ = .25), ESR (one-way ANOVA; mild, 18.5 ± 16.5; moderate, 21.9 ± 16.9; severe, 21.1 ± 17.2, $p$ = .79), and CRP (one-way ANOVA; mild, 6.8 ± 11.9; moderate 6.0 ± 6.3; severe, 5.9 ± 6.0; $p$ = .95). However, a chi-square test for independence led us to reject the null hypothesis that pain intensity and pathology are independent variables ($p$ < .001). Regarding the interviewer influence, one interviewer performed 83% of the interviews, whilst the remaining are similarly distributed by the other two interviewers. Again, a chi-square test for independence led us to reject the null hypothesis that the reported pain intensity and who the interviewer was are independent variables ($p$ < .001).

*4.4. Classification results*

The classification results for the two machine learning models with each of the 6 features sets are presented in Table 3, as well as the performance of the ZRB. The best performance was obtained when using the technique of EF, specifically with the SVM model, which had an increase of 35% performance compared to the baseline. This result is further explored in Table 4, where the confusion matrix for the SVM model with EF features is shown, again, in comparison to the ZRB.

*4.5. Influence of confounding variables on language*

We then observed how the SVM model, with the initial set of EF features (also denoted EF + SVM), behaved if we discarded all features for which we did not have sufficient evidence to reject correlation with demographic and clinical parameters. The results are shown in Table 5. Note that certain demographic or clinical parameters, for which no language features were found to have evidence of correlation, such as age, time since diagnostics, time since the start of pain reports, ESR, and CRP, are omitted. Thus, for a given demographic or clinical parameter, e.g., gender, the model EF + SVM (all features), predicts female and male samples with a weighted $F_1$ score of .48 and .73, respectively. However, when we discarded the features for which we did not have sufficient evidence to reject correlation with gender (select features), and trained the model, it then predicted female and male samples with a weighted $F_1$ score of .52 (+.04) and .71 (−.02), respectively. Moreover, the aggregate score increased from .58, the original performance of the EF + SVM model, to .59 (+.01).

The performance results in Table 5 show that the highest combined gain is obtained when dropping the features that correlate with whether the participant is professionally active or not. The confusion matrix of this score is shown in Table 6.

*4.6. Language differences between pain intensity categories*

By analysing the EF + SVM model (trained with select features so that the dependency, on whether the participant is professionally active or not, is minimized), whose scores are shown in Table 5 and the confusion matrix in Table 6, we inferred what allowed the model to distinguish between participants with mild, moderate, and severe pain.

In the LOOV setting we trained $n = 65$ models on $n - 1$ samples, each of which selecting $k \in [2,20]$ features, according to the experimental setup previously described. In total, 117 unique features were chosen and all of them belonged to the TF-IDF set. Because we preemptively tagged each individual word in the TF-IDF set with its POS tag, we could observe the syntactic distribution of these features, as shown in Fig. 1: the most important features (in terms of relative count), are nouns, verbs, adjectives, and adverbs, in this order. Fig. 2 shows the average importance given by each pain intensity category to the features (words) of each of these POS tags. Participants with severe pain focused the most on nouns and adverbs, moderate pain participants focused the most on adverbs, and participants with mild pain focused the most on verbs. All pain categories focused the least on adjectives.

Table 7 shows the topmost weighted words (on average), for each of the most relevant POS tags in Fig. 1. Fig. 3 shows the average importance of each of the words in Table 7, for each pain intensity category. All scores belong to a small scale; therefore, focus will be given on the most striking differences. Regarding nouns, events during the year and at home were most relevant to severe pain participants, whilst nightly events were most relevant to participants reporting moderate pain (Fig. 3a). In accordance with Fig. 2, there is not a noun most relevant to mild pain participants. Regarding verbs, there is a similar importance distribution for all pain categories, in general (Fig. 3b). However, mild pain participants were singled-out by their focus on the verbs "to worry", and "to lock". Regarding adjectives, all pain categories give a similarly low importance to these words, in fact, the lowest (Fig. 3c). Finally, regarding adverbs, participants feeling moderate pain seem to use more descriptive terms than other pain categories (Fig. 3d).

## 5. Discussion (1500 words)

To the best of our knowledge, this is the first research work that attempts to estimate current pain intensity from language, specifically from chronic pain patient narratives. The best model (EF + SVM with select features) performed 39.5% better than the (zero rate) baseline, showing that participants with severe pain focused the most on nouns and adverbs, moderate pain participants focused the most on adverbs, and participants with mild pain focused the most on verbs. Adjectives were focused the least by all pain categories.

Our sample size compares to other similar works [8,3,33,29,11,34,4], even though it is not very large, due to participant availability and time constraints. This means that our sample of chronic pain patients has limited representativeness (participants in our sample are mostly above 50 years of age, female, with a basic education level, and with more than 10 years of reported pain.). We attempted to circumvent this by extensively testing the null hypothesis of independence between all captured demographic and clinical parameters, and the target pain intensity categories. Indeed, there is not sufficient statistical evidence in our sample to reject the possibility of correlation between the reported pain intensity and gender, education level, whether the participant was professionally active or not, pathology, and who the interviewer was. Interestingly, in our sample of participants, there was no significant statistical difference in ESR and CRP for the 3 pain intensity categories, which are both clinical assessments to detect inflammation, commonly correlated with pain.

Regarding data preprocessing, our pipeline is substantially based on semi-automatic and manual processes, which are particularly expensive and susceptible to human errors. To overcome this, we manually validated a sub-sample of intermediate and final preprocessing results. We identified a few errors, but deemed unnecessary to validate the whole dataset, due to the low error rate. We identified some lemmatization and POS tagging errors in later stages, but, again, these were deemed insignificant.

The classification results were obtained in a LOOV setting, which is well-suited for small datasets, such as ours. According to the experimental setup, a model is only deemed useful for the task if it performs better than the ZRB. Since both DT and SVM models perform below the baseline when using the POS TFIDF or topic distribution features, we can state that these features, by themselves, are not useful for the task of multi-class pain intensity classification. The best results were obtained with the SVM model, using EF features. The corresponding confusion matrix allows us to conclude that there is confusion with mild pain, and that the most difficulty arises when distinguishing mild and severe pain participants. Considering that there might exist language dependencies and modulation, ultimately affecting prediction generalizability (e.g., if the model heavily uses feature *X* for prediction, and that feature is highly correlated with gender, then the model is effectively using gender for prediction, and not language), we removed all language features for which we had sufficient evidence of correlation with one of these parameters, thus, broadening the model's generalizability. The performance results show that there is a performance gain in ignoring features for which there is evidence of correlation with the participant's gender, education level, and whether was professionally active (according to the aggregate score, when compared to the EF + SVM model performance). The highest combined gain was obtained when dropping the features that correlate with whether the participant was professionally active or not. The comparison of the corresponding confusion matrices suggests that the improvement is specifically in differentiating moderate from mild pain. However, the difficulty in differentiating severe from mild pain remains. Indeed, our results show that our model (specifically, EF + SVM with select features) is useful for the multi-class pain intensity classification problem, performing 39.5% better than the baseline (which itself deterministically predicts the most prevalent target class). Moreover, our model accurately classifies 92% of mild, 42% of moderate, and 13% of severe pain cases. These results function as a proof-of-concept for what could be achieved with larger datasets.

The results indicate that, for our sample of chronic pain participants, the most important words that allow to distinguish between pain intensity categories are, in order, nouns, verbs, adjectives, and adverbs. Indeed, nouns allow for the best distinction of severe pain, adverbs for moderate pain, and verbs for mild pain. This is expected, since the remaining POS, such as pronouns, determinants, and prepositions, carry very little semantic value, instead playing a more syntactic role in the construction of sentences. Most words directly relate to some dimension of the experience of chronic pain, e.g., treatment, periods of time, feelings and sensations, suffering, enduring, and surpassing pain. These are the words found to be most relevant for the task of pain intensity classification. More specifically, nouns seem mostly used to refer to treatments, time periods, and parts of the body. Verbs seem to refer mostly to feeling pain and fearing its outcomes. Multiple adjectives are used, of which we highlight the importance of social activity, nervousness, tranquility, and pain-related sensations, such as dryness. Finally, adverbs are used to specify other words, such as adjectives and verbs. Regarding time periods, mild pain associates more importance to the month, while moderate and severe pain associate more importance to the night and to the year, respectively (although not all, some of these differences are indeed striking). It also tells us that the verb "to suffer" is mostly important to severe pain, whilst "to endure" and "to go away" are mostly important to moderate and mild pain, respectively. As expected, all pain categories give high importance to the verb "to feel". Our feature extraction did not capture negatives, thus not allowing us to judge the use of these specific verbs. It may very well be the case that mild pain participants are stating that they are not worried about their pain (given the importance shown for the verb "to worry"), but we cannot know that for a fact with our model. The same applies to the verb "to go away", as it could be used to state that the pain is going away, or not. The adjective social (assumingly used in the context of social activities, as prompted by the interview question) is important for moderate and severe pain, but not to mild pain, most likely due to the impact caused by the pain on the day-to-day life. Finally, moderate pain participants seem to use more descriptive adverbs than patients with mild and severe pains, which suggests a need to detail a possibly ambiguous experience, one that is clearly neither mild, nor severe. Indeed, our model takes advantage of these differences to perform the task, however they are not sufficient for perfect classification.

With this work, we have demonstrated a proof-of-concept for the analysis of the language of chronic pain for current pain intensity estimation. We have shown that focus on specific words/themes is especially correlated with specific pain intensity categories. Importantly, our approach does not limit chronic pain patients to close-ended questions and answers, allowing them to freely discuss the concerns that they find most relevant at the time of reporting. This inverts the standard assessment methods by letting the patient guide the narrative to their important factors, instead of the other way around, which could help reduce the impact of social biases in self-reports and not force the patient to translate the personal experience of pain into impersonal molds.

In conclusion, with this work, we have explored the applicability of language, more specifically of chronic pain narratives from the personal experience of pain, to predict the current pain intensity. We described the sample population, experimented with a diverse experimental setup, and explored the results obtained beyond prediction performance, observing the language features that allowed to better distinguish mild, moderate, and severe pains. We have also identified the main limitations and strengths of our work.

Future research should focus on expanding the sample size so that demographic and clinical parameters may be properly tested for independence with the target variable, and possibly controlled during experimentation. We also suggest a focus on automating the whole preprocessing pipeline to reduce costs and have better control over the expected results, especially if the dataset is larger.

## 7. Conflict of interest statement

The authors have no conflicts of interest to declare.

## 8. Acknowledgements

Diogo A.P. Nunes is supported by a scholarship granted by Fundação para a Ciência e Tecnologia (FCT), with reference 2021.06759.BD. This work was supported by Portuguese national funds through FCT, with reference UIDB/50021/2020.

Due to privacy concerns and the ethics protocol followed in this work, our data is not publicly available.

**Figure Legends**

**Figure 1.** Title: Percentage of used features that belong to each Part-of-Speech tag. Description: POS distribution of the TF-IDF features used for pain intensity classification by the best performing model. NOUN refers to nouns, VERB refers to verbs, ADJ refers to adjectives, ADV refers to adverbs, and OTHERS includes the remaining POS tags identified by STRING (13, in total).

**Figure 2.** Title: Average weight given by participants in each pain class to each Part-of-Speech tag of used features. Description: Average weight given to the features (words) in each of the POS tags shown in Fig. 1, per pain intensity category.

**Figure 3a.** Title: Average weight given by participants in each pain class to each noun (from the top-10). Description: Average word importance for each pain intensity class, as given by the TF-IDF set of features. Words are separated per assigned POS and correspond to the words shown in Table 7. Words are solely presented in English.

**Figure 3b.** Title: Average weight given by participants in each pain class to each verb (from the top-10). Description: Average word importance for each pain intensity class, as given by the TF-IDF set of features. Words are separated per assigned POS and correspond to the words shown in Table 7. Words are solely presented in English.

**Figure 3c.** Title: Average weight given by participants in each pain class to each adjective (from the top-10). Description: Average word importance for each pain intensity class, as given by the TF-IDF set of features. Words are separated per assigned POS and correspond to the words shown in Table 7. Words are solely presented in English.

**Figure 3d.** Title: Average weight given by participants in each pain class to each adverb (from the top-10). Description: Average word importance for each pain intensity class, as given by the TF-IDF set of features. Words are separated per assigned POS and correspond to the words shown in Table 7. Words are solely presented in English.

**Figure 1.**

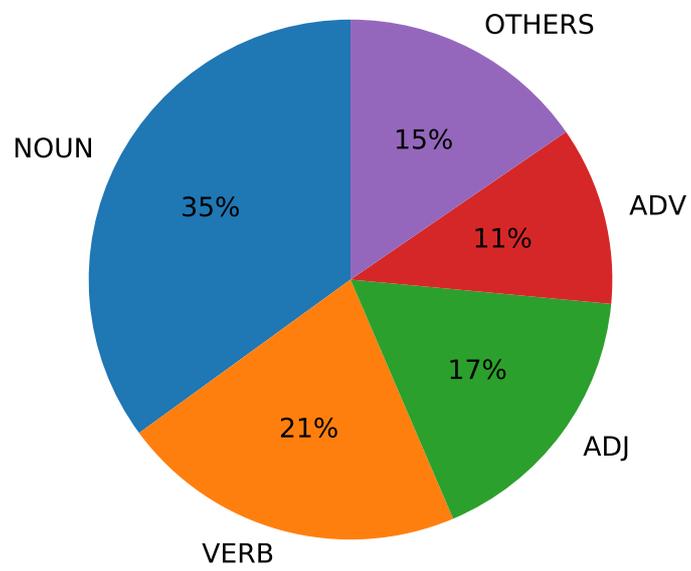

**Figure 2.**

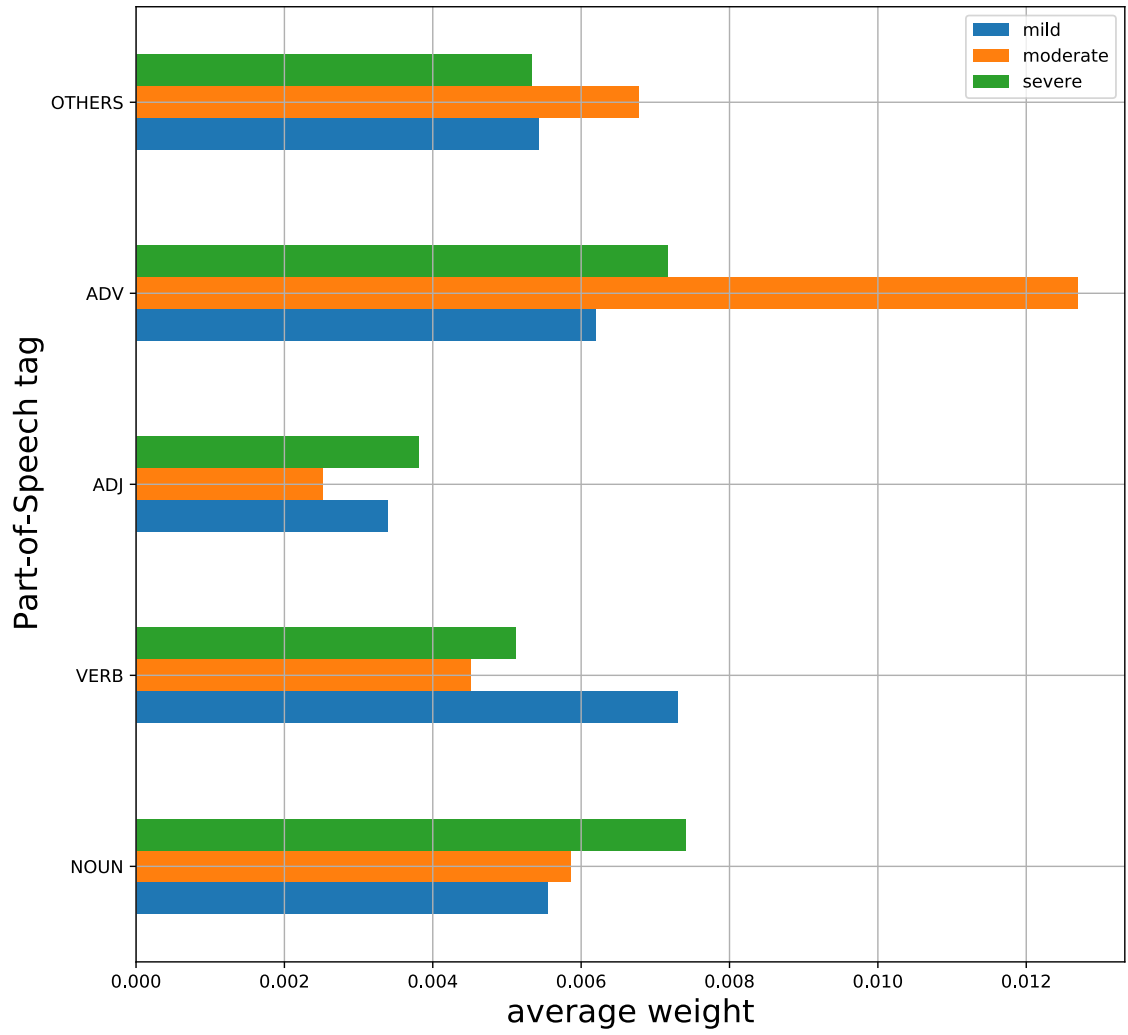

**Figure 3a.**

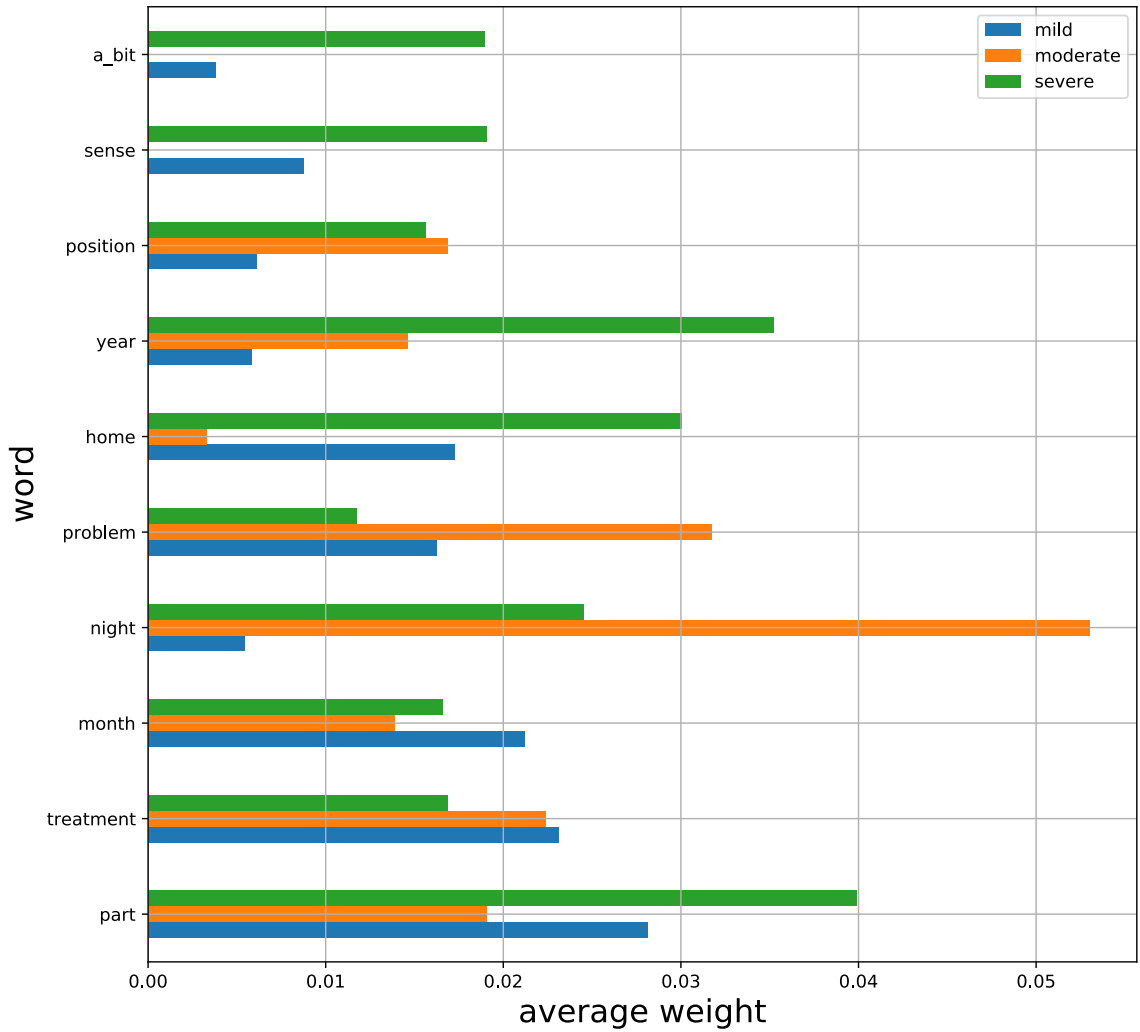

**Figure 3b.**

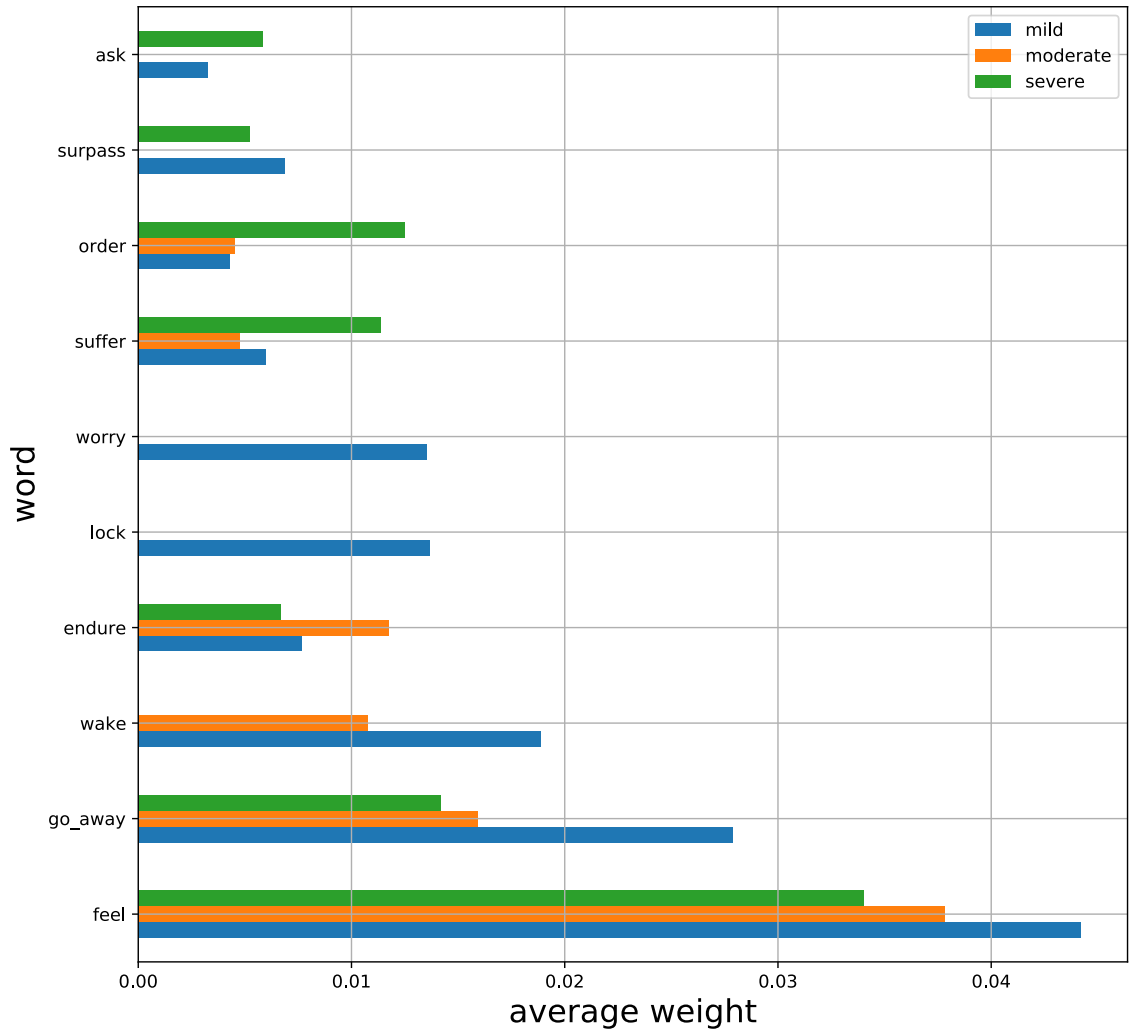

**Figure 3c.**

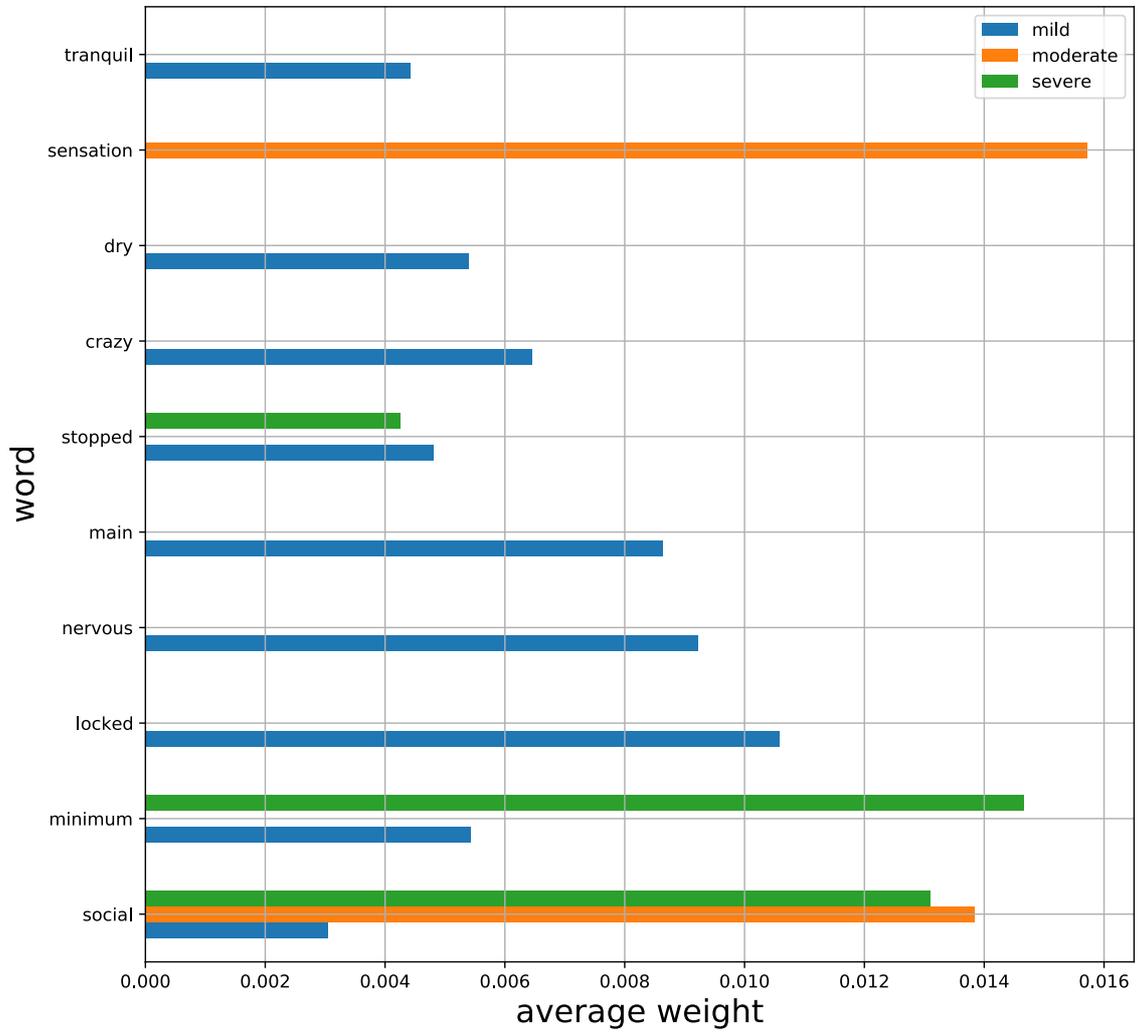

**Figure 3d.**

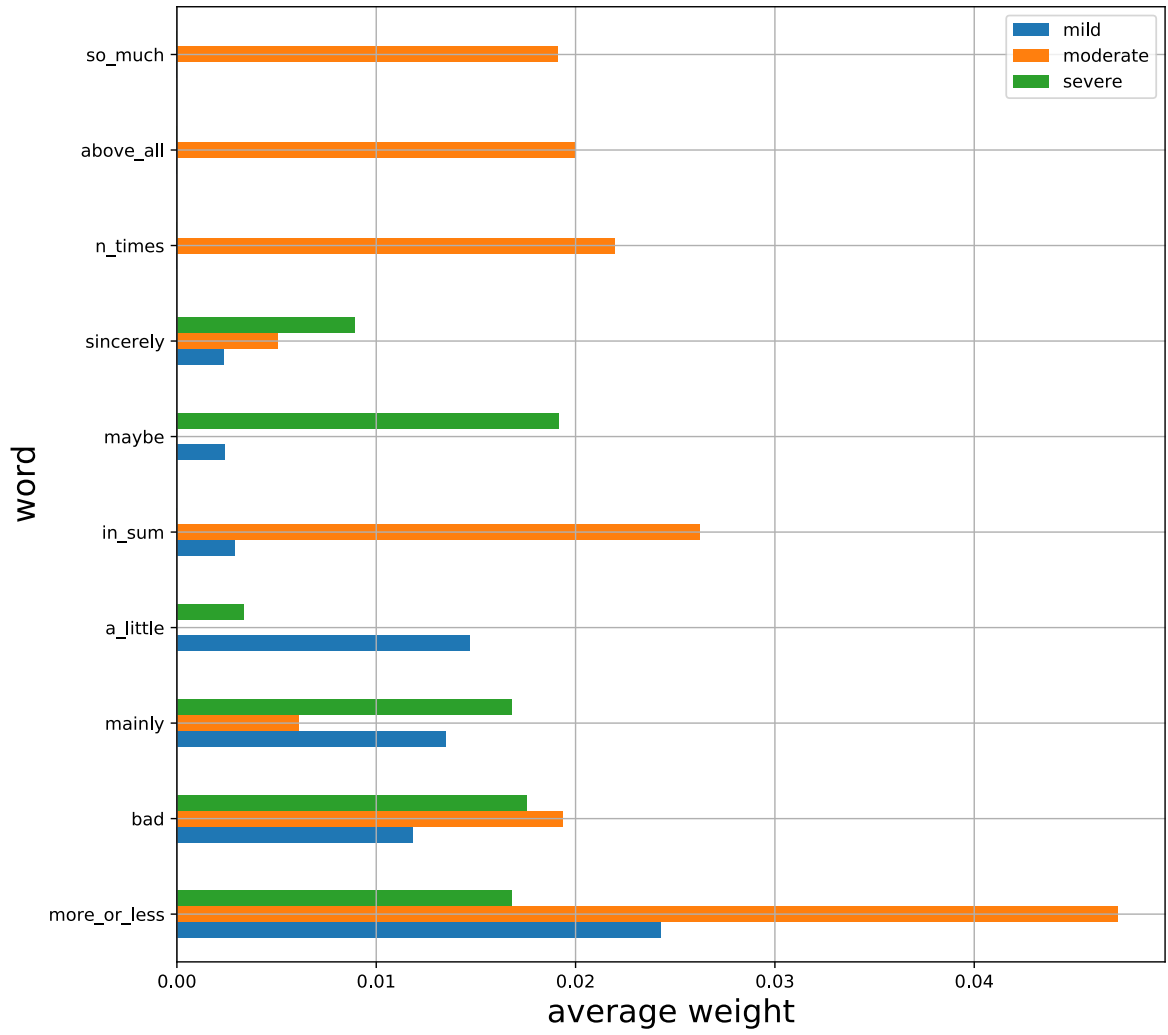

**Table 1.**

Title: Data collection interview questions, topics, and reasoning.

| Number | Question | Topic | Reasoning |
|---|---|---|---|
| 1. | Where does it hurt? | Location | The physical location, perceived sensations, and intensity of pain are known to influence the overall experience, as well as affect the psychosocial dimension of pain [10]. |
| 2. | How would you describe your pain? How do you feel it/which sensations does it cause? | Sensation | |
| 3. | How has pain intensity evolved in the past month? | Intensity | |
| 4. | How would you consider pain to affect your day-to-day, namely, your physical, professional, and social activities and your emotional state? | Impact | Psychology of pain research has shown that there are various psychological and social aspects that influence one's perception and the consequent experience of pain [9]. Thus, for the psychosocial dimension of the experience of chronic pain, we included questions regarding the impact, perceived cause, and belief of pain. |
| 5. | What do you believe to be the cause of your pain? | Cause | |
| 6. | How would you say your pain has evolved, considering the current treatment? | Treatment effect | It has been shown that the patient's satisfaction with the current treatment is correlated with its adherence and consequent outcome [9]. |
| 7. | How do you expect your pain to develop in the coming months? | Belief, expectations | See reasoning of questions 4 and 5. |

Description: The questions presented here are translated from Portuguese. The interview is composed of 7 questions, each focusing on a specific topic regarding the personal experience of chronic pain. For each question, the corresponding topic and reasoning is shown. Question numbering indicates the interview script order.

Legend: N/A

**Table 2.**

Title: Distribution of demographic and clinical variables for each pain intensity category.

| Pain | Age | Gender | Edu. level | Professionally active | Interviewer | Total |
| --- | --- | --- | --- | --- | --- | --- |
| mild | 55.8 ± 13.0 | 21×F, 17×M | 27×B, 11×H | 20×A, 18×NA | 4×I1, 29×I2, 5×I3 | 38 |
| moderate | 54.7 ± 12.5 | 6×F, 6×M | 10×B, 2×H | 7×A, 5×NA | 2×I1, 10×I2 | 12 |
| severe | 59.1 ± 10.9 | 13×F, 2×M | 12×B, 3×H | 8×A, 7×NA | 3×I1, 11×I2, 1×I3 | 15 |
| **Pain** | **Time diagn.** | **Time pain** | **ESR** | **CRP** | **Pathology** | **Total** |
| mild | 11.3 ± 9.5 | 14.6 ± 11.2 | 18.5 ± 16.5 | 6.8 ± 11.9 | 20×RA, 16×S, 2×O, 1×PA | 38 |
| moderate | 11.9 ± 9.1 | 14.8 ± 10.7 | 21.9 ± 16.9 | 6.0 ± 6.3 | 4×RA, 7×S, 3×O | 12 |
| severe | 15.4 ± 8.6 | 20.1 ± 9.0 | 5.9 ± 6.0 | 5.9 ± 6.0 | 5×RA, 9×S, 1×PA | 15 |

Description: Note that 2 participants have comorbid osteoarthritis and rheumatoid arthritis, and 1 participant has comorbid osteoarthritis and spondyloarthritis. Variables Age, Time diagnostics, and Time pain are reported in years.

Legend: F = female, M = male, B = basic education, H = high-school education, A = professionally active, NA = not professionally active (retired or on sick leave), I1 = interviewer 1, I2 = interviewer 2, I3 = interviewer 3, RA = rheumatoid arthritis, S = spondyloarthritis, O = osteoarthritis, PA = psoriatic arthritis, ESR = Erythrocyte Sedimentation Rate, CRP = C-Reactive Protein.

**Table 3.**

Title: Weighted $F_1$ scores for each type of features.

| Feature set | ZRB | DT | SVM |
|---|---|---|---|
| Verbosity | .43 | .44 | .43 |
| TF-IDF | .43 | .55 | .51 |
| POS TF-IDF | .43 | .36 | .40 |
| Topic distribution | .43 | .39 | .39 |
| Early Fusion (EF) | .43 | .56 | **.58** |
| Late Fusion (LF) | .43 | .46 | .40 |

Description: The $F_1$ score ranges from 0 to 1, where 1 is the best possible score (i.e., perfect classification).

Legend: Zero Rate Baseline (ZRB), Decision Tree Classifier (DT), and Support Vector Machine (SVM). Up-arrow means that the higher the score, the better.

**Table 4.**

Title: Confusion matrix for the Early Fusion + Support Vector Machine model predictions.

|  |  | Predicted label | | |
| --- | --- | --- | --- | --- |
|  |  | mild | moderate | severe |
| Real label | mild | **.92** (1.0) | .00 (.00) | .08 (.00) |
|  | moderate | .58 (1.0) | **.33** (.00) | .09 (.00) |
|  | severe | .73 (1.0) | .13 (.00) | **.13** (.00) |

Description: The confusion matrix for the Zero Rate Baseline is shown in parentheses for comparison.

Legend: N/A.

**Table 5.**

Title: Weighted $F_1$ scores using Early Fusion + Support Vector Machine model, discriminated by demographic and clinical categories.

|  | Gender | | Edu. Level | | Pathology | | | | Prof. active | | Interviewers | | |
|---|---|---|---|---|---|---|---|---|---|---|---|---|---|
|  | F | M | B | H | AR | S | O | PA | A | NA | I1 | I2 | I3 |
| ZRB | .36 | .55 | .56 | .39 | .44 | .43 | .33 | .33 | .56 | .23 | .27 | .43 | .76 |
| All features | .48 | **.73** | .56 | .57 | **.67** | **.49** | **.68** | .00 | .51 | .62 | .38 | **.58** | .76 |
| Select features | **.52** | .71 | **.56** | **.59** | .61 | .41 | .23 | **.33** | **.55** | **.62** | **.39** | .54 | **.76** |
| Agg. score | .59 | | **.60** | | .49 | | | | **.60** | | .54 | | |

Description: Weighted $F_1$ scores using Early Fusion + Support Vector Machine model, discriminated by demographic and clinical categories, using either all features, or just select features, according to a statistical independence test (the aggregate score compares directly to the using Early Fusion + Support Vector Machine model). The discriminated Zero Rate Baseline scores are shown for baseline comparison. The $F_1$ score ranges from 0 to 1, where 1 is the best possible score (i.e., perfect classification).

Legend: ZRN = Zero Rate Baseline, F = female, M = male, B = basic education, H = high-school education, A = professionally active, NA = not professionally active (retired or on sick leave), I1 = interviewer 1, I2 = interviewer 2, I3 = interviewer 3, RA = rheumatoid arthritis, S = spondyloarthritis, O = osteoarthritis, PA = psoriatic arthritis.

**Table 6.**

Title: Confusion matrix for the Early Fusion + Support Vector Machine model predictions with select features.

|  |  | Predicted label | | |
| --- | --- | --- | --- | --- |
|  |  | mild | moderate | severe |
| Real label | mild | **.92** (1.0) | .00 (.00) | .08 (.00) |
|  | moderate | .50 (1.0) | **.42** (.00) | .08 (.00) |
|  | severe | .73 (1.0) | .13 (.00) | **.13** (.00) |

Description: The confusion matrix for the Zero Rate Baseline is shown in parentheses for comparison.

Legend: N/A.

**Table 7.**

Title: List of the top-10 most weighted words (on average) of the most relevant Part-of-Speech tags.

| POS | Top-10 weighted words |
|---|---|
| Noun | parte (*part*), tratamento (*treatment*), mês (*month*), noite (*night*), problema (*problem*), casa (*home*), ano (*year*), posição (*position*), sentido (*sense*), um_bocado (*a_bit*) |
| Verb | sentir (*feel*), passar (*go_away*), acordar (*wake*), suportar (*endure*), prender (*lock*), preocupar (*worry*), sofrer (*suffer*), mandar (*order*), ultrapassar (*surpass*), pedir (*ask*) |
| Adjective | social (*social*), mínimo (*minimum*), preso (*locked*), nervoso (*nervous*), principal (*main*), parado (*stopped*), maluco (*crazy*), seco (*dry*), sensação (*sensation*), tranquilo (*tranquil*) |
| Adverb | mais_ou_menos (*more_or_less*), mal (*bad*), principalmente (*mainly*), pouco (*a_little*), na_totalidade (*in_sum*), talvez (*maybe*), sinceramente (*sincerely*), n_vezes (*n_times*), sobretudo (*above_all*), tanto (*so_much*) |

Description: List of the top-10 most weighted words (on average) of the most relevant POS tags, as suggested by the results shown in Fig.1. English translation is shown in parenthesis.

Legend: POS = Part-of-Speech